\documentclass{article}

\usepackage{arxiv}

\usepackage[utf8]{inputenc} 
\usepackage[T1]{fontenc}    
\usepackage{hyperref}       
\usepackage{url}            
\usepackage{booktabs}       
\usepackage{amsfonts}       
\usepackage{nicefrac}       
\usepackage{microtype}      
\usepackage{lipsum}		
\usepackage{graphicx}
\usepackage{natbib}
\usepackage{doi}
\usepackage{xcolor}         
\usepackage{algorithmic}
\usepackage{algorithm}
\usepackage{subcaption}
\usepackage{multirow}
\usepackage{enumitem}

\title{Neuroevolution deep learning architecture search for estimation of river surface elevation from photogrammetric Digital Surface Models}

\date{} 					

\author{ 
    \href{https://orcid.org/0000-0003-0769-7755}{\includegraphics[scale=0.06]{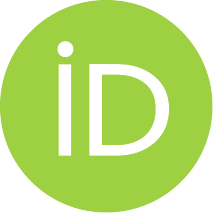}\hspace{1mm}
    Radosław Szostak}\\
	AGH UST\\
	\texttt{rszostak@agh.edu.pl} \\
	\And
	\href{https://orcid.org/0000-0001-9357-9231}{\includegraphics[scale=0.06]{orcid.pdf}\hspace{1mm}
    Marcin Pietroń}\\
    AGH UST\\
    \texttt{pietron@agh.edu.pl} \\
    \And
    \href{https://orcid.org/0000-0002-0594-9376}{\includegraphics[scale=0.06]{orcid.pdf}\hspace{1mm}
    Mirosław Zimnoch} \\
    AGH UST\\
    \texttt{zimnoch@agh.edu.pl} \\
    \And
    \href{https://orcid.org/0000-0001-6563-8249}{\includegraphics[scale=0.06]{orcid.pdf}\hspace{1mm}
    Przemysław Wachniew} \\
    AGH UST\\
    \texttt{wachniew@agh.edu.pl} \\
    \And
    \href{https://orcid.org/0000-0001-5526-0908}{\includegraphics[scale=0.06]{orcid.pdf}\hspace{1mm}
    Paweł Ćwiąkała} \\
    AGH UST\\
    \texttt{pawelcwi@agh.edu.pl} \\
    \And
    \href{https://orcid.org/0000-0003-0607-0432}{\includegraphics[scale=0.06]{orcid.pdf}\hspace{1mm}
    Edyta Puniach} \\
    AGH UST\\
    \texttt{epuniach@agh.edu.pl} \\
}

\renewcommand{\shorttitle}{\textit{arXiv} Template}

\hypersetup{
pdftitle={A template for the arxiv style},
pdfsubject={q-bio.NC, q-bio.QM},
pdfauthor={David S.~Hippocampus, Elias D.~Striatum},
pdfkeywords={deep learning, geoscience, neuroevolution, climate change},
}

\begin{document}
\maketitle

\begin{abstract}
    Development of the new methods of surface water observation is crucial in the perspective of increasingly frequent extreme hydrological events related to global warming and increasing demand for water. Orthophotos and digital surface models (DSMs) obtained using UAV photogrammetry can be used to determine the Water Surface Elevation (WSE) of a river. However, this task is difficult due to disturbances of the water surface on DSMs caused by limitations of photogrammetric algorithms.
    In this study, machine learning was used to extract a WSE value from disturbed photogrammetric data. A brand new dataset has been prepared specifically for this purpose by hydrology and photogrammetry experts.
    The new method is an important step toward automating water surface level measurements with high spatial and temporal resolution. Such data can be used to validate and calibrate of hydrological, hydraulic and hydrodynamic models making hydrological forecasts more accurate, in particular predicting extreme and dangerous events such as floods or droughts.
    For our knowledge this is the first approach in which dataset was created for this purpose and deep learning models were used for this task. Additionally, neuroevolution algorithm was set to explore different architectures to find local optimal models and non-gradient search was performed to fine-tune the model parameters. The achieved results have better accuracy compared to manual methods of determining WSE from photogrammetric DSMs.
    
\end{abstract}


\section{Introduction}

Reports from international organizations indicate increasingly significant problems with earths water resources. The global demand for freshwater continues to increase at rate 1\% per year since 1980s driven by population growth and socioeconomic changes. Simultaneously, the increase in evaporation caused increasing temperatures leads to a decrease in streamflow volumes in many areas of the world, which already suffer from water scarcity problems. Climate warming is also responsible for globally increased frequency of extreme hydrologic conditions. More intense and frequent precipitation events increase the flood risk as well as heatwaves are becoming more common and last longer, resulting in more severe droughts (\cite{unesco2020}, \cite{ipcc2014}). Achieving socioeconomic and environmental sustainability under such challenging conditions will require the use of monitoring tools that will facilitate the management of the water resources. Traditional surface water management practices are primarily based on data collected from networks of in situ hydrometric gauges. Point measurements do not provide sufficient spatial resolution to comprehensively characterize river networks, and many developing regions lack them altogether. Moreover, the decline of existing measurement networks is being observed all over the world (\cite{Lawford2013}). Remote sensing methods are considered as a solution to cover data gaps specific to point measurement networks (\cite{McCabe2017}). A leading example of remote sensing is measurements made from satellites. However, due to too low spatial resolution, satellite data is suitable only for studying the largest rivers. E.g. SWOT mission allows only observation of rivers of width greater than 50-100 m (\cite{Pavelsky2014}). Small surface streams of the first and second order (according to Strahler's classification (\cite{Strahler1957}) constitute 70\%-80\% of the length of all rivers in the world. Small streams play a significant role in hydrologic systems and provide an ecosystem for living organisms  (\cite{Wohl2017}). Despite their high importance, modern methods for their observation are still lacking. In this regard, measurement techniques based on Unmanned Aerial Systems (UASs) are promising in many key aspects, as they are characterized by high spatial and temporal resolution, simple and fast deployment, and the ability to be used in inaccessible locations (\cite{VlezNicols2021}). Spatialy distributed Water Surface Elevation (WSE) measurements are highly important, as they are used for validation and calibration of hydrologic, hydraulic or hydrodynamic models to make hydrological forecasts, including predicting dangerous events such as floods and droughts (\cite{Langhammer2017}, \cite{Tarpanelli2013}, \cite{AsadzadehJarihani2013}, \cite{Domeneghetti2016}, \cite{Montesarchio2014})

Photogrammetric Structure from Motion (SfM) algorithms are able to generate Orthophotos and Digital Surface Models (DSMs) of terrain based on multiple aerial photographs. Photogrammetric DSMs are precise in determining the elevation of solid surfaces to within a few cm (\cite{Oudraogo2014}, \cite{Bhler2017}). However, they do not correctly represent the water surface. This is related to the fact that general principle of SfM algorithms is based on automatic search for a distinguishable and static terrain points that appear in several images showing these points from different perspectives. The surface of the water lacks such points as it is uniform, transparent and in motion. Due to water transparency, DSMs created using SfM algorithms typically indicate pixel elevations below the actual water surface level. For very clear and shallow streams, photogrammetric DSMs represent the river bottom (\cite{Kasvi2019}). For opaque waters, photogrammetric DSMs are disturbed by artifacts caused by water uniformity (lack of distinguishable photogrammetric key-points). \cite{Woodget2014}, \cite{Javernick2014}  and \cite{Pai2017} demonstrated that it is possible to read the WSE from photogrammetric DSM at shorelines ("water-edge") where river is very shallow, so there are no undesirable effects associated with light penetration below the water surface. However, \cite{Bandini2020} proved that this method gives satisfactory results only for unvegetated and smoothly sloping shorelines where the boundary line between water and land is easy to define. For this reason, this method is not suitable for universal automation. Table \ref{statistc_of_methods} shows the RMSE errors of existing Remote Sensing methods for small rivers WSE measurements.

\begin{table}[]
\centering
\caption{\centering{Remote sensing small river WSE measurement error comparison. RMSE values taken from: UAV -- \cite{Bandini2020}, AIRSWOT -- \cite{Altenau2017}}}
\begin{tabular}{|c|c|}
\hline
\textbf{Method} & \textbf{RMSE (m)}  \\ \hline
UAV RADAR     &  0.03  \\ \hline
AIRSWOT    &   0.09  \\ \hline
UAV SfM DSM centerline & 0.164  \\ \hline
UAV SfM point cloud & 0.180 \\ \hline
UAV LIDAR point cloud   &  0.22    \\ \hline
UAV SfM DSM "water-edge" & 0.450  \\ \hline
\end{tabular}
\label{statistc_of_methods}
\end{table}

The aim of this work was to develop a new automatic method based on deep neural networks allowing estimation of small rivers WSE from photogrammetric DSMs and Orthophotos with an accuracy outperforming previous methods based on manual analysis of photogrammetric data.

\section{Dataset}
\subsection{Dataset structure}
In this research, models were trained using brand new dataset, created specifically for the purpose of this work. It consist of 260 samples. Each sample corresponds to a 10 by 10 meter area that encloses river water and nearshore land. Dataset is divided into a training and testing subset at a ratio of 8:2. Dataset is available to download at \url{https://zenodo.org/record/5257183} (\cite{szostak_dataset_v1}).
Every sample includes the data detailed below.
\begin{itemize}
    \item \textbf{Photogrammetric orthophoto}. True color image represented as a $3\times256\times256$ array (3 channel image of $256\times256$ pixels).
    \item \textbf{Photogrammetric DSM}. Corresponds to the area presented on the orthophoto. Contains disturbed water surfaces elevations. Stored as a $256\times256$ array containing elevations of pixels expressed in m~MSL.
    \item \textbf{WSE}. Ground truth Water Surface Elevation as single value expressed in m~MSL.
\end{itemize}
\begin{figure}[h!]
  \centering
  \begin{subfigure}[t]{0.2\linewidth}
    \includegraphics[height=3cm]{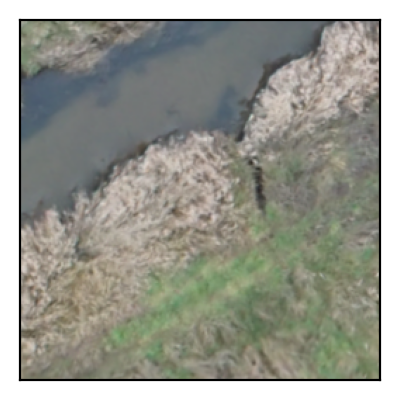}
    \caption{}
  \end{subfigure}
  \begin{subfigure}[t]{0.2\linewidth}
    \includegraphics[height=3cm]{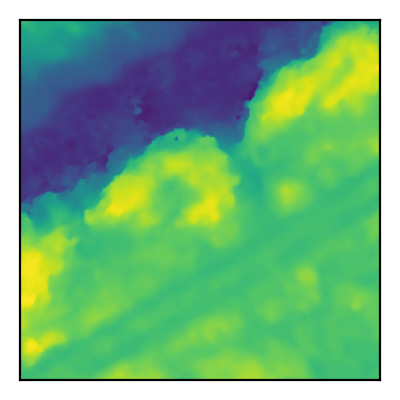}
    \caption{}
  \end{subfigure}
  \begin{subfigure}[t]{20pt}
    \includegraphics[height=3cm]{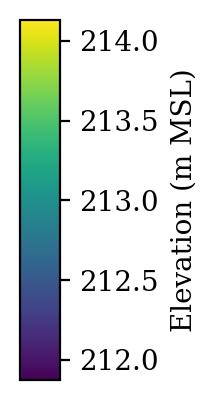}
  \end{subfigure}
  \caption{Visualisation of geospatial data from single dataset sample. (a)~--~photogrammetric orthophoto, (b)~--~raw photogrammetric DSM with water surface disturbances.}
  \label{fig:coffee3}
\end{figure}

\subsection{Dataset preparation}
The photogrammetric data and WSE observations needed to create the dataset was obtained during surveys in two study areas:
\begin{enumerate}
\item in a ca. 700m stretch of the Kocinka river located near Grodzisko village (Silesian Voivodeship, Poland) on December 19, 2020 and July 13, 2021. This stretch had a WS width of ca. 2m and was overhung by sparse deciduous trees that had no leaves due to the winter season. The banks as well as the riverbed were overgrown with rushes that protruded above the water surface. The banks are steeply sloping at angles of ca. 50\textdegree\ to 90\textdegree\ relative to the water surface. There are marshes nearby, with river water flowing into them in places.
\item in a ca. 700m stretch of the Kocinka river located near Mykanów village (Silesian Voivodeship, Poland) on December 19, 2020 and July 13, 2021. This stretch had a WS width of ca. 3m and was overhung by sparse deciduous trees. Grasses growing on the banks slightly overhang the water surface. The banks are steeply sloping at angles of ca. 50\textdegree\ to 90\textdegree\ relative to the water surface.
\end{enumerate}

In addition, the dataset was supplemented with data from surveys made by
\cite{bandini_data} over a ca. 2.3km stretch of the river Åmose Å (Denmark) on November 21, 2018. See the cited publication for details on this river case study: \cite{Bandini2020}.

During survey campaigns, photogrammetric measurement were conducted over the river area. The aerial photos were taken from a DJI S900 UAV using a Sony ILCE a6000 camera with a Voigtlander SUPER WIDE HELIAR VM 15 mm f/4.5 lens. The flight altitude was approximately 77m AGL, resulting in a 20mm terrain pixel. Photos front overlap was 80\%, and side overlap was 60\%. Alongside the drone measurements, measurements of ground cotrol points were made using RTK GPS receiver. These were used to embed the photogrammetric rasters in the geographic reference system. RTK GPS receivers were also used for reference point measurements of WSE. They were made along the river every approximately 10-20 meters at both banks. Raster files of orthophotos and digital surface models of the investigated areas were generated using Agisoft Metashape photogrammetric software. 

The measurement campaigns resulted in raster files with width and height of several tens of thousands of pixels, each representing an area of about 40ha. For use in machine learning algorithms, they were divided into 256px x 256px images representing sub-areas of actual size in the field of 10m x 10m. Point ground truth measurements of river water surrface elevation were interpolated using the IDW method. Interpolated values were assigned to individual samples based on geospatial location.

\section{Deep Learning solution}
\subsection{Feature scaling}
Input data is subjected to feature scaling before it is fed into the model. DSMs values were standardized according to the equation $DSM'=\frac{DSM-\overline{DSM}}{2\sigma}$, where $DSM'$ -- standardized sample DSM 2D array with values centered around 0, $DSM$ -- raw sample DSM 2D array with values expressed in m~MSL, $\overline{DSM}$ -- mean value of single subjected $DSM$ array, $\sigma$ -- standard deviation of DSM array pixel values from the entire dataset. This method of standardization has two clear advantages. Firstly, by subtracting the average value of a single subjected sample, standardized DSMs are always centered around zero, so the algorithm sees no difference between samples of the rivers located at regions of different altitudes. The actual water level information is recovered during inverse standardization. Secondly, dividing all samples by the same sigma value of entire dataset, ensures that all standardized samples are scaled equally. It was experimentally found that multiplying the denominator by 2 results in better model accuracy, compared to standardization that does not include this factor. Orthophotos were standardized using Imagenet (\cite{Deng2009}) dataset mean and standard deviation according to the equation $ORT'=\frac{ORT-\mu}{\sigma}$, where $ORT'$ -- standardized 3-channel orthophoto RGB image 3D array with values centered around 0, $ORT$ -- 3-channel orthophoto RGB image 3D array represented with values from the range [0,1], $\mu = [0.485, 0.456, 0.406]$ -- 1D vector containing mean values of each of RGB channels from Imagenet dataset, $\sigma = [0.229, 0.224, 0.225]$ -- 1D vector containing standard deviation values of each of RGB channels from Imagenet dataset.

\subsection{Models}
\label{wse_model}
The model used to create the supervised learning algorithm for determining a single WSE value is based on the VGG-16 architecture (\cite{simonyan2015deep}). Several variations of this model have been tested:
\begin{enumerate}[leftmargin=*,noitemsep,topsep=-5pt,label=\textbf{\alph*}]
    \item \label{level_base_model} \textbf{VGG-16 Base Model}. VGG-16 originally used for image classification was modified to perform single floating point value prediction. The changes made to this model are: i) the input size of the model is 4x256x256. It is a four-channel image in which the first channel contains the DSM and the other three channels are RGB orthophoto channels. ii) After a series of convolution layers, a linear transformation of the array data to a single value was applied. No activation function was used on the model output to obtain a floating point value.
    \item \textbf{Multiresolution VGG-16}. VGG-16 Base Model (\ref{level_base_model}) enhanced with multi resolution achieved by concatenation of scaled four-input channels to the output of each max pooling layer.
    \item \textbf{VGG-16 with four conv blocks} - VGG-16 without last three conv layers with changed activation function to PReLU.
    \item \textbf{VGG-16 with three conv blocks} - VGG-16 without last six conv layers with ReLU6.
    \item \textbf{VGG-16 with five blocks} - VGG-16 with whole feature extractor with number of channels 256 changed to 148).
    \item \textbf{fine tuned VGG-16} - best pretrained VGG-16 fine tuned by running neuroevolution with weights mutation.
    \item \textbf{ResNeST50} - Resnet with split attention model 
    \item \textbf{VisionTransformer} - Vision Transformer model with different number of heads and depth parameter
    \item \textbf{EfficientNet} - b0 and b1 EfficientNet architectures
\end{enumerate}




\subsection{Architecture search}
Many of recent machine learning works has focused on solutions in which neural network weights are trained through variants of stochastic gradient descent. An alternative approach comes from the field of neuroevolution, which harnesses evolutionary algorithms to optimize neural networks, inspired by the fact that natural brains themselves are the products of an evolutionary process (\cite{faber2021}, \cite{sceneNet}, \cite{NEAT}, \cite{neuro_evolving}, \cite{neuro_image}, \cite{neuroevol_overview}). In presented work we have set up a neuroevolution based algorithm which can run search through different architectures and modifications of few baseline models VGG, ResNet, ResneST, EfficientNet, VisionTransformer (\ref{alg:neuro}). The size of population in our experiments was 16  (eq.\ref{eq:population}, eq. \ref{eq:model}). The number of iterations is in range from 20 to 40. The population is set of the models with different initial random weights ((eq.\ref{eq:population}, eq. \ref{eq:model})). 
Equation \ref{eq:population} describes the definition of the model population where $\theta$ is a weight tensor.

\begin{equation}
P_M = \{F_{\Theta}^{i}, \Theta = \{\theta_{0}, \theta_{1},...,\theta_{N}\} \land i \in \{1, 2, ..., population\_size\}\} 
\label{eq:population}
\end{equation}

\begin{equation}
F_{\Theta}^{i}(X) = {f_{\theta_N}^{i}(f_{\theta_{N-1}}^{i}...(f_{\theta_{0}}^{i}(X)))}
\label{eq:model}
\end{equation}




We define the crossover as a function that has two parent models and an id of the layer as input parameters (Equation ( \ref{eq:crossover}), alg.\ref{alg:neuro}, line 6). It generates two new children models (Equation (\ref{eq:crossover_function}, alg. \ref{alg:crossover}, line 11)). 

\begin{equation}
c:Parents \times layer\_id \rightarrow Children
\label{eq:crossover}
\end{equation}

\begin{equation}
c(F_{\Theta_{M}}^{i}, F_{\Theta_{N}}^{j}, l\_id) \rightarrow (F_{\Theta_{M'}}^{i'}, F_{\Theta_{N'}}^{j'}) 
\label{eq:crossover_function}
\end{equation}

\begin{equation}
f_{\theta_{l\_id}}^{i'} = f_{\theta_{l\_id}}^{j}
\label{eq:change_1}
\end{equation}

\begin{equation}
f_{\theta_{l\_id}}^{j'} = f_{\theta_{l\_id}}^{i}
\label{eq:change_2}
\end{equation}

The crossover can exchange between parents activation functions, convolutional, or fully connected layers (eq.\ref{eq:change_1}, eq.\ref{eq:change_2}, alg.\ref{alg:crossover}, line 3 and line 6). The last option is to exchange hyperparameters between parents, such as  learning rate and weight decay (alg.\ref{alg:crossover}, line 9). We define mutation as a function with a model, layer id, and layer features to be mutated as input parameters (Equation \ref{eq:mutation}, alg.\ref{alg:mutation}). In this case, the feature and $layer\_id$ determine what the mutation operator will do. It can change the activation function or parameters of convolutional or fully connected layers (depth) (\ref{alg:mutation}, line 3 and line 6). The other option is to change the length of the model (adding or removing a layer) or update the hyperparameters. The result is a new modified model (Equation (\ref{eq:mutation_func})).

\begin{equation}
m:Parent \times layer\_id \rightarrow Child \label{eq:mutation}
\end{equation}

\begin{equation}
m(F_{\Theta_{M}}^{i}, l\_id) \rightarrow F_{\Theta_{N}}^{i'}
\label{eq:mutation_func}
\end{equation}

\begin{equation}
f_{\theta_{l\_id}}^{i'} = depth(f_{\theta_{l\_id}}^{i}) \pm \beta \cdot \Delta_{depth} 
\label{eq:mutation_func}
\end{equation}

Finally, evolution process is finished, non-gradient fine tuning is invoked. Fine tuning is performed as weight mutation on a pretrained model (eq.\ref{eq:mutation_func_}, eq.\ref{eq:mutation_func__}, alg.\ref{alg:fine}). The additional input parameter for the non gradient fine tuning is the scaling factor for computing the value by which the weights can be modified (eq.\ref{eq:mutation_func___}).
The constant percentage of weights in each weight tensor are selected for fine tuning process. The selection process is done by binary $\alpha$ tensor. The $\alpha$ tensor is dynamic and can be different in each fine tuning iteration.

\begin{equation}
m:Parent \times scale \rightarrow Child
\label{eq:mutation_func_}
\end{equation}

\begin{equation}
F_{\Theta'}^{i'} = m(F_{\Theta}^{i}), \Theta' = \{\theta'_{0}, \theta'_{1},...,\theta'_{N}\}
\label{eq:mutation_func__}
\end{equation}

\begin{equation}
w'_{ij} = w_{ij} \pm \alpha_{ij} \cdot\frac{max(abs(\theta_{i}))}{scale_i}
\label{eq:mutation_func___}
\end{equation}





\begin{algorithm}
\begin{algorithmic}[1]
\REQUIRE{$P_{S}$ - population size, $M_B$ - base model, $N_{iter}$ - number of iterations}
\STATE{$P$ $\gets$ generate population with $P_{S}$ size from $M_{B}$}
\FOR{$i$ \textbf{in} $N_{iter}$}
\FOR{$s$ \textbf{in} $P$}
\STATE{$p$ $\gets$ get random parent from $P$}
\STATE{$l$ $\gets$ choose layer for exchange}
\STATE{$p1$, $p2$ $\gets$ crossover($s$, $p$, $l$)}
\STATE{$P$ $\gets$ $P$ $\cup$ $p1$ $\cup$ $p2$}
\STATE{$s'$ $\gets$ mutation($s$)}
\STATE{$P$ $\gets$ $P$ $\cup$ $s'$} 
\ENDFOR
\STATE{compute accuracy for all models from $P$}
\STATE{$P$ $\gets$ take $P_{S}$ best models from $P$}
\ENDFOR
\end{algorithmic}
\caption{Neuroevolution approach}
\label{alg:neuro}
\end{algorithm}

\begin{algorithm}
\begin{algorithmic}[1]
\REQUIRE{$p1$ - parent 1, $p2$ - parent 2, $l$ - layer for an exchange in crossover}
\STATE{$L$ $\gets$ get layer $l$ from parent $p1$} 
\IF{$L$ \textbf{is} activation $type$}
\STATE{$c1$, $c2$ $\gets$ exchange activation $L$ layer between $p1$ and $p2$}
\ENDIF
\IF{$L$ \textbf{is} CONV \textbf{or} FC}
\STATE{$c1$, $c2$ $\gets$ exchange layer $L$ between $p1$ and $p2$}
\ENDIF
\IF{$L$ \textbf{is} None}
\STATE{$c1$, $c2$ $\gets$ exchange hyperparameters between $p1$ and $p2$}
\ENDIF
\STATE{Return $c1$ and $c2$}
\end{algorithmic}
\caption{Crossover operator}
\label{alg:crossover}
\end{algorithm}

\begin{algorithm}
\begin{algorithmic}[1]
\REQUIRE{$M$ - model, $l$ - layer id}
\STATE{$L$ $\gets$ get $l$ layer from model $M$}
\IF{$L$ \textbf{is} activation $type$}
\STATE{$M'$ $\gets$ change activation function of $L$ in $M$}
\ENDIF
\IF{$L$ \textbf{is} CONV or FC}
\STATE{$M'$ $\gets$ change parameters of layer $L$ in $M$ //in depth, out depth, filter size, padding etc.}
\ENDIF
\IF{$L$ \textbf{is} None}
\STATE{$M'$ $\gets$ change optimizer of $M$}
\ENDIF
\STATE{Return $M'$}
\end{algorithmic}
\caption{Mutation operator}
\label{alg:mutation}
\end{algorithm}

\begin{algorithm}
\begin{algorithmic}[1]
\REQUIRE{$M_{p}$ - pretrained model, $P_{S}$ - population size of fine tuning, $N_{g}$ - Number of generations in the fine tuning, $p$ - percentage of weights to be perturbated, $\alpha$ - scaling factor for magnitude perturbation}
\STATE{$P$ $\gets$ generate population with $P_{S}$ size from $M_{p}$}
\STATE{$generation = 0$}
\WHILE{$generation < N_{g}$} 
\FOR{$s$ \textbf{in} $P$}
\STATE{$s'$ $\gets$ mutate randomly $p$ percentage of weights in $s$ using \ref{eq:mutation_func___}}
\STATE{calculate the fitness of $s'$}
\ENDFOR
\STATE{P $\gets$ choose the $P_{S}$ best solutions and include them in the new population}
\STATE{$generation \gets generation + 1$}
\ENDWHILE
\STATE{Return $P$}
\end{algorithmic}
\caption{Fine tuning}
\label{alg:fine}
\end{algorithm}





\section{Results and future works}

\begin{table}[]
\centering
\caption{Results on 260 images.}
\begin{tabular}{|c|c|}
\hline
\textbf{Model} & \textbf{water level estimation error}  \\ \hline
vgg 4 blocks with LeakyReLU (84,137,258,497)    &  11.18   \\ \hline
vgg 3 blocks with ReLU  &  10.44 \\  \hline
vgg 5 blocks with LeakyReLU (66,146,176,222,222)    &   10.01 \\ \hline
\textbf{fine tuned vgg 5 with LeakyReLU}  & \textbf{9.89} \\ \hline
VGG-16 Base Model  & 11.69 \\ \hline
Resnet18 Model  (42, 193, 293, 579) & 14.13 \\ \hline
Multiresolution VGG-16  &  10.50 \\ \hline
\textbf{ResneST50 with PReLU} (88,192,198,202) [3,5,5,4] & \textbf{8.58} \\  \hline
EfficientNet-b0 & 16.12 \\ \hline
EfficientNet-b1 & 13.75 \\ \hline
\textbf{VisionTransformer (depth=8, heads=11)} & \textbf{9.99}\\ \hline

\end{tabular}
\label{results}
\end{table}


The results are shown in Tab.~\ref{results}. The models listed in table are those which were set manually (VGG-16  Base Model, Multiresolution VGG-16 Base Model) and other which were generated by neuroevolution search. In parentheses there are combinations of the number of channels for successive blocks inside the models. In the rows where these numbers are not given, the number of channels in the blocks is the same as in the original version of the model. It is shown that neuroevolution search improves accuracy of water level prediction (e.g. ResneST with modified number of sub-blocks - [3,5,5,4], VisionTransformer (depth=8, heads=11)). Our fine tuning approach decrease further the prediction error in VGG model. It is worth to mention that our best deep learning models outperform other manual methods of determining WSE from photogrammetric DSMs and are close to the accuracy of the more complicated and expensive AirSWOT method (Tab. \ref{statistc_of_methods}).
The future work will concentrate on further model exploration using more sophisticated genetic operators which can modify number of sub-blocks, run models clustering, hyperparameter optimization (\cite{faber2021}) and automatically apply multiresolution options. Also Bayesian estimation of uncertainty and models sensitivity analysis will be performed (\cite{gal2016dropout}).

\bibliographystyle{unsrtnat}
\bibliography{references}

\end{document}